\documentclass[sigconf,nonacm]{acmart}

\setcopyright{none}
\copyrightyear{2026}
\acmYear{2026}

\usepackage{amsmath}
\usepackage{graphicx}
\usepackage{tikz}
\usetikzlibrary{arrows.meta, positioning, fit, backgrounds, shapes.geometric, calc}
\usepackage{multirow}
\usepackage{makecell}
\usepackage{tcolorbox}
\tcbuselibrary{breakable}
\usepackage{array}
\usepackage{ragged2e}

\usepackage[capitalize]{cleveref}
\crefname{section}{Sec.}{Secs.}
\Crefname{section}{Section}{Sections}
\crefname{appendix}{Appendix}{Appendices}
\Crefname{appendix}{Appendix}{Appendices}
\Crefname{table}{Table}{Tables}
\crefname{table}{Tab.}{Tabs.}
\Crefname{figure}{Figure}{Figures}
\crefname{figure}{Fig.}{Figs.}

\usepackage{xspace}
\makeatletter
\DeclareRobustCommand\onedot{\futurelet\@let@token\@onedot}
\def\@onedot{\ifx\@let@token.\else.\null\fi\xspace}

\makeatother

\newcommand{\todo}[1]{}

\newcommand{\BackboneGemma}{Gemma 4 26B A4B}
\newcommand{\BackboneOss}{gpt-oss-20b}
\newcommand{\SettingGemmaLow}{\BackboneGemma{} low}
\newcommand{\SettingGemmaMedium}{\BackboneGemma{} medium}
\newcommand{\SettingOssLow}{\BackboneOss{} low}
\newcommand{\SettingOssMedium}{\BackboneOss{} medium}
\newcommand{\SettingHeaderGemmaLow}{\makecell[r]{\BackboneGemma{}\\low}}
\newcommand{\SettingHeaderGemmaMedium}{\makecell[r]{\BackboneGemma{}\\medium}}
\newcommand{\SettingHeaderOssLow}{\makecell[r]{\BackboneOss{}\\low}}
\newcommand{\SettingHeaderOssMedium}{\makecell[r]{\BackboneOss{}\\medium}}
\newcommand{\SettingCellGemmaLow}{\makecell[l]{\BackboneGemma{}\\low}}
\newcommand{\SettingCellGemmaMedium}{\makecell[l]{\BackboneGemma{}\\medium}}
\newcommand{\SettingCellOssLow}{\makecell[l]{\BackboneOss{}\\low}}
\newcommand{\SettingCellOssMedium}{\makecell[l]{\BackboneOss{}\\medium}}

\begin{document}

\title{Total Recall at What Cost? Benchmarking the Serving Cost of Agentic
  Memory Systems}

\author{Natchanon Pollertlam}
\email{natchanon.p@brickstech.co}
\affiliation{%
  \institution{Bricks Technology}
  \country{Thailand}
}

\author{Witchayut Kornsuwannawit}
\email{witchayut.k@brickstech.co}
\affiliation{%
  \institution{Bricks Technology}
  \country{Thailand}
}

\renewcommand{\shortauthors}{Pollertlam and Kornsuwannawit}

\begin{abstract}
Long-running conversational agents increasingly rely on a memory system to
avoid resending the whole conversation each turn, yet how much that costs to
serve has received little systematic benchmarking. We compare three memory
systems (Mem0, Hindsight, and Mastra Observational Memory) against two reference
strategies --- a fixed-size rolling window and resubmitting the full
transcript --- across two backbones and conversations of up to 400 turns,
pairing every cost measurement with answer accuracy on 665 LoCoMo questions.
First, a memory system's serving cost cannot be predicted from conversation
length and message size alone: a regression that tracks the two reference
strategies closely misses the memory systems by $18$--$69\%$, their cost driven
instead by internal memory behavior. Second, a break-even analysis shows that
whether --- and when --- a memory system becomes cheaper to serve than the full
transcript is highly sensitive to the system and the backbone, from the first
tens of turns for the cheapest to never within 400 turns for the most
expensive. Third, no system wins on both axes: accuracy spans $21$--$54\%$, and
the backbone choice drives cost as much as the memory system does.
\end{abstract}



\maketitle

\section{Introduction}
\label{sec:intro}

Conversational agents built on large language models (LLMs) have moved from
single-session demonstrations into deployments that span weeks or months of
interaction. To stay coherent across sessions, such an agent must reuse
information established in earlier turns. Two strategies address this. The
first resubmits the full prior transcript into the context window of a
long-context LLM on every turn, relying on the model to attend over all past
interaction. The second builds a dedicated \emph{memory system} that distills
past interaction into compact records --- extracted facts, summaries, or
knowledge-graph entries --- and retrieves only the relevant subset at query
time~\cite{lewis2020retrieval,packer2024memgptllmsoperatingsystems,chhikara2025mem0buildingproductionreadyai}.
The second strategy has since been developed into a variety of architectures.

As these agents scale, deployment cost becomes a first-order concern alongside
accuracy. Commercial LLM APIs price requests by token
count~\cite{openai2025pricing,anthropic2025pricing}, so, because the
history grows with every turn, the per-turn cost of resubmitting it rises without
bound as a session continues. Memory systems are adopted, in large part, to
escape this growth: by replacing an ever-larger transcript with a small
retrieved payload, they promise a per-turn cost that stays roughly flat as a
conversation lengthens. This framing, however, overlooks the cost of the
memory system itself. Modern memory systems are not passive stores; they run
their own LLM pipelines --- extracting facts at ingest, embedding and
retrieving them, and in some designs periodically reflecting over accumulated
state to consolidate it. Each stage incurs its own billable model calls. Yet
evaluation of memory systems concentrates almost entirely on accuracy and
recall~\cite{maharana2024evaluatinglongtermconversationalmemory,wu2025longmemevalbenchmarkingchatassistants,jiang2025personamemv2personalizedintelligencelearning},
and what cost evidence exists appears as isolated token-savings or
latency figures reported by individual systems under their own conditions. No
study has measured the serving cost of memory systems across systems, under a
common backbone and pricing, against a transparent baseline.

As a result, a practitioner choosing a memory system today cannot answer a
basic question: what does it cost to serve, and does that cost actually beat
resubmitting the transcript? A memory system's per-turn cost is shaped by its
own internal pipeline behavior, and whether it is economical further depends on
how long the conversation runs and which backbone model serves it. Without a
controlled, cross-system measurement, the cost case for memory systems rests on
assumption rather than evidence.

In this study, we benchmark the serving cost of agentic memory systems. We
measure three memory systems ---
Mem0~\cite{chhikara2025mem0buildingproductionreadyai},
Hindsight~\cite{latimer2025hindsight2020buildingagent}, and Mastra
Observational Memory~\cite{barnes2026observationalmemory} --- against two
reference strategies that bracket the cost surface: a fixed-size rolling window
and full-transcript resubmission. We run every system across two backbone
models at two reasoning-effort settings, on synthetic conversations of up to
400 turns, and pair each cost measurement with answer accuracy on the LoCoMo
benchmark~\cite{maharana2024evaluatinglongtermconversationalmemory} so that
cost and accuracy are read from a single matched configuration. Our
contributions are:

\begin{itemize}
    \item A controlled benchmark of the serving cost of three agentic memory
    systems, measured against two transparent reference strategies: a
    rolling window and full-transcript resubmission.
    \item A per-turn cost model that fits message size and
    conversation depth as two independent terms and is validated on held-out
    workloads; it predicts the reference strategies but not the memory systems,
    whose cost is driven by internal memory state.
    \item A break-even analysis that identifies when a memory system becomes
    cheaper to serve than resubmitting the full transcript.
    \item A joint cost--accuracy comparison on LoCoMo across memory systems and
    backbone models.
\end{itemize}

\section{Related Work}
\label{sec:related}

\noindent
\textbf{Memory systems for conversational agents.}
Retrieval-augmented generation~\cite{lewis2020retrieval} is an early form of
memory augmentation, prepending retrieved document chunks to the prompt to
ground generation in external knowledge. MemGPT~\cite{packer2024memgptllmsoperatingsystems}
draws an analogy between LLM context management and operating-system virtual
memory, paging facts between an in-context working memory and external
long-term storage. Mem0~\cite{chhikara2025mem0buildingproductionreadyai}
extracts atomic, flat-typed facts from each turn and stores them in a vector
database, retrieving the top-$k$ at query time. Later systems pursue richer
representations, including interlinked memory
notes~\cite{xu2025amemagenticmemoryllm}, temporally-aware knowledge
graphs~\cite{rasmussen2025zeptemporalknowledgegraph}, and durative summaries of
temporally continuous facts~\cite{su2026dialoguetimetemporalsemantic}. The three systems we benchmark span
this design space: Mem0 represents flat extract-and-retrieve,
Hindsight~\cite{latimer2025hindsight2020buildingagent} adds a
retain--recall--reflect memory ingestion pipeline, and Mastra
Observational Memory~\cite{barnes2026observationalmemory} runs an
observer--reflector--actor loop with threshold-triggered consolidation.
Crucially, these architectures differ not only in what they store but in
\emph{how they spend compute} --- a fixed-size fact extraction, a retrieval payload 
that grows as the conversation lengthens, or a threshold-triggered memory-consolidation 
pass --- so they cannot be assumed to share a single cost profile. Prior work characterizes
these systems by their representations and recall accuracy; we characterize them by serving cost.

\noindent
\textbf{Benchmarking memory and long context.}
Several benchmarks evaluate how well a system recalls extended interaction.
LoCoMo~\cite{maharana2024evaluatinglongtermconversationalmemory} provides
multi-session dialogues whose questions span single-hop, multi-hop,
temporal, and open-domain categories;
LongMemEval~\cite{wu2025longmemevalbenchmarkingchatassistants} tests information
extraction, multi-session reasoning, temporal reasoning, knowledge updates, and
abstention; and PersonaMem~\cite{jiang2025personamemv2personalizedintelligencelearning}
probes persona consistency across questions. Such benchmarks have also been
used to weigh memory systems against long-context inference directly, as
expanding context windows raise the question of whether retrieval remains
necessary~\cite{khalusova2024ragvslc,rengifo2025longercontext}, even though
long-context models attend unevenly across a long prompt~\cite{liu2024lost}.
That debate, however, has been conducted almost entirely on accuracy: these
benchmarks measure what a system recalls, not what it costs to serve, and the
serving cost of a memory system is left uninstrumented. Our work pairs accuracy on LoCoMo
with a matched cost measurement so that the two are directly comparable.

\noindent
\textbf{Cost and efficiency of LLM inference.}
Per-token API pricing makes inference cost a central production concern, and a
range of techniques target it. Prompt caching reuses the precomputed key-value
states of a shared input prefix~\cite{gim2024promptcachemodularattention}, and
providers discount cached input tokens
steeply~\cite{openai2024promptcaching,anthropic2024promptcaching}; prompt
compression instead shortens the input itself~\cite{jiang2023longllmlingua}.
Such techniques optimize a \emph{single} inference path. A memory system, by
contrast, is a multi-stage pipeline whose total billed cost compounds an ingest
stage, a retrieval stage, and an answer stage, each issuing its own model
calls. The compositional cost of such a pipeline --- and the conversation
length at which it undercuts simply resubmitting the transcript --- has not
been characterized. We provide that characterization: a controlled
serving-cost benchmark of memory systems against transparent floor and ceiling
baselines, with a break-even analysis and a matched accuracy comparison.

\section{Methodology}
\label{sec:method}

We evaluate three memory systems in two separate benchmarks that share a
common backbone, reasoning-effort, and embedding configuration: a
\textbf{cost} benchmark, measuring billable serving cost as a function of
conversation length and turn size, and an \textbf{accuracy} benchmark,
measuring answer correctness on a stratified subset of a multi-session
conversational QA benchmark. \Cref{sec:joint-framing} combines them into a
single cost-per-correct-answer statistic.

\subsection{Memory Systems}
\label{sec:systems}

We benchmark three memory systems: \textbf{Mem0}~\cite{chhikara2025mem0buildingproductionreadyai},
\textbf{Hindsight}~\cite{latimer2025hindsight2020buildingagent}, and
\textbf{Mastra Observational Memory} (Mastra OM)~\cite{barnes2026observationalmemory}.
Per-system pipeline configuration is tabulated in \Cref{tab:app-systems}.

We additionally evaluate two reference strategies that bracket the cost
surface:
\begin{description}
  \item[Rolling window (floor).] Each turn sees only the last 10 turns;
  nothing is stored persistently --- the cheapest possible strategy.
  \item[Full history (ceiling).] The entire transcript is resubmitted every
  turn. Every memory system's break-even point (\Cref{sec:breakeven}) is
  measured against this ceiling.
\end{description}

\subsection{Backbone Models and Reasoning Levels}
\label{sec:backbone}

All systems are evaluated under the same two backbones,
\textbf{gpt-oss-20b}~\cite{openai2025gptoss120bgptoss20bmodel} and
\textbf{Gemma 4 26B A4B}~\cite{googledeepmind2026gemma4}, each at two
reasoning-effort settings (\emph{low}, \emph{medium}). Embeddings use
\texttt{pplx-embed-v1-0.6b}
\cite{eslami2026diffusionpretraineddensecontextualembeddings}; per-token
pricing and OpenRouter provider routing are given in \Cref{app:config}.

\subsection{Cost Benchmark Design}
\label{sec:cost-design}

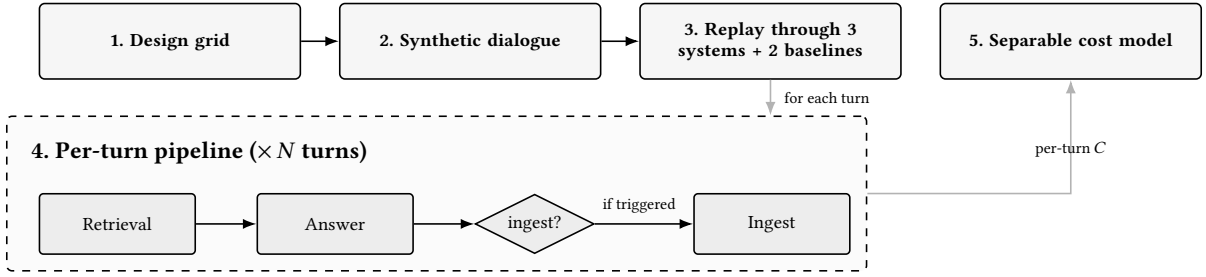
\begin{figure*}[t]
\centering
\begin{tikzpicture}[
  font=\footnotesize,
  >={Latex[length=1.8mm]},
  stage/.style={draw, semithick, rounded corners=2pt, fill=gray!7,
                text width=3.1cm, align=center, inner sep=5pt,
                minimum height=1.0cm},
  pnode/.style={draw, semithick, rounded corners=1.5pt, fill=gray!14,
                text width=1.85cm, align=center, inner sep=3pt,
                minimum height=0.8cm},
  gate/.style={draw, semithick, fill=gray!14, diamond, aspect=2.0,
               align=center, inner sep=2pt},
  arr/.style={->, semithick},
  link/.style={->, semithick, draw=gray!60},
]

\node[stage] (grid) {\bfseries 1.~Design grid};
\node[stage, right=0.5cm of grid] (gen) {\bfseries 2.~Synthetic dialogue};
\node[stage, right=0.5cm of gen] (replay)
  {\bfseries 3.~Replay through 3 systems + 2 baselines};
\node[stage, right=0.5cm of replay] (model)
  {\bfseries 5.~Separable cost model};

\draw[arr] (grid) -- (gen);
\draw[arr] (gen) -- (replay);

\node[pnode, below=1.5cm of grid.south west, anchor=north west] (retr)
  {Retrieval};
\node[pnode, right=0.8cm of retr] (answ) {Answer};
\node[gate, right=0.8cm of answ] (trig) {ingest?};
\node[pnode, right=1.3cm of trig] (ingt) {Ingest};

\draw[arr] (retr) -- (answ);
\draw[arr] (answ) -- (trig);
\draw[arr] (trig) -- node[pos=0.45, above=1pt, font=\scriptsize] {if triggered} (ingt);

\node[font=\bfseries, anchor=south west] (pttitle)
  at ([xshift=-6pt, yshift=8pt] retr.north west)
  {4.~Per-turn pipeline ($\times\,N$ turns)};

\begin{scope}[on background layer]
  \node[draw, dashed, semithick, rounded corners=2pt, fill=gray!3,
        inner sep=6pt, fit=(pttitle)(retr)(answ)(trig)(ingt)]
        (ptbox) {};
\end{scope}

\draw[link] (replay.south) -- node[midway, right=2pt, font=\scriptsize] {for each turn}
  (replay.south |- ptbox.north);
\draw[link] (ptbox.east) -| node[pos=0.6, above=2pt, font=\scriptsize] {per-turn $C$}
  (model.south);

\end{tikzpicture}
\caption{Overview of the cost benchmark. (1)~We sample five $(N,L)$ cells:
the four corners and the center of a grid over conversation length $N$ and
per-turn token size $L$. (2)~For each $(N,L,\mathrm{seed})$, an LLM generates
a two-speaker conversation that is cached and reused unchanged, so every
system sees the same input. (3)~Each cached dialogue is replayed through
three memory systems and two reference baselines. (4)~Within a run, every
turn passes through retrieval, answer, and an ingest gate that decides
whether to flush the buffered turns; the input tokens billed across these
three stages give the per-turn cost~$C$. (5)~We then fit a log-log cost
model with separate $N$ and $L$ terms, one fit per (system, model),
validated by leave-one-out cross-validation with bootstrap confidence
intervals.}
\Description{A two-tier flow diagram. The top row shows four boxed stages
left-to-right: ``1. Design grid'', ``2. Synthetic dialogue'', ``3. Replay
through three memory systems and two baselines'', and ``5. Separable cost
model''. Arrows connect each stage to the next. Below stage 3, a
dashed-outline group labeled ``4.
Per-turn pipeline (x N turns)'' contains four sub-nodes in sequence:
``Retrieval'', ``Answer'', a diamond gate ``ingest?'', and ``Ingest''. An
arrow from stage 3 enters this per-turn group; an arrow exits it carrying
``per-turn C'' into stage 5.}
\label{fig:pipeline}
\end{figure*}

\Cref{fig:pipeline} gives an end-to-end overview of the cost benchmark, from
the design grid through the fitted cost model; the remainder of this
subsection details each stage.

\paragraph{Grid.}
For each (system, model) pair we sample five $(N, L)$ cells: the four corners
and the center of a grid over conversation length $N$ and per-turn token
size $L$. Each cell is run eight times with different random seeds. The exact
$(N, L)$ values, including extra cells for the baselines and Mastra OM, are
listed in \Cref{app:bench}.

\paragraph{Synthetic dialogue generation.}
Each cost-benchmark dialogue is generated synthetically by an LLM: for each
$(N, L, \text{seed})$ the model produces a two-speaker conversation of $N$
turns at roughly $L$ tokens per turn, prompted to keep concrete personal
detail --- names, dates, preferences, plans. Over-length turns are trimmed and
under-length turns padded so turn lengths stay close to $L$; each finished
dialogue is cached so it is identical across repeats and across systems.
Generating dialogues lets us fill every $(N, L)$ cell exactly. The generation
prompt is reproduced in \Cref{app:prompts}.

\paragraph{Cost model.}
\label{sec:cost-forms}
For a turn at depth $t$ in a conversation whose messages average $L$ tokens,
let $C$ be the total LLM input tokens billed across ingest, retrieval, and
answer. We model $C$ rather than dollar cost: dollar cost follows from the
per-token rates of \Cref{app:config}, and output tokens, which do not grow
with depth, we track separately through the $\gamma$ diagnostic below.

Conversation depth and message size scale cost differently, so we fit them
as two separate log-log terms rather than as one cumulative-content
predictor ($N{\cdot}L$):
\begin{equation}
\log(C{+}1) = a + p\,\log(L{+}1) + q\,\log(t{+}1),
\label{eq:multivariate}
\end{equation}
where $p$ is how fast cost grows with message size and $q$ how fast it grows
with depth: $q\!\approx\!0$ is the signature of a bounded context window;
$p\!\approx\!q\!\approx\!1$ is cumulative content. We fit this form once per
(system, model). Alongside it we report three token-accounting diagnostics:
$\gamma$ (output-to-input ratio), $\zeta_\mathrm{ans}$ (answer-stage
reasoning-to-output ratio), and $\zeta_\mathrm{ing}$ (ingest-stage
reasoning-to-output ratio).

\paragraph{Held-out validation.}
We test whether each fitted model generalizes by leave-one-cell-out
cross-validation~\cite{stone1974cross}: for every (system, model) we drop one $(N, L)$ cell, refit
\Cref{eq:multivariate} on the rest, and predict the dropped cell's mean
per-turn cost. The error measure, LOOCV-MAPE, is the mean absolute percentage
error over the held-out cells. We use it as a diagnostic, not a pass/fail bar. A low value means message
size and depth alone account for the system's cost. A high value is itself a
finding: it shows that the system's cost is driven by internal memory state
that $(L, t)$ cannot capture (\Cref{sec:loocv}).

\paragraph{Confidence intervals.}
Confidence intervals for $p$ and $q$ come from a cluster
bootstrap~\cite{efron1979bootstrap,cameron2008bootstrap} with
$B{=}1{,}000$ resamples. A \emph{run} is one end-to-end execution of a
synthetic dialogue through the system; each cell contributes 8 runs at
different seeds. The bootstrap resamples whole runs, not individual turns, because turns within
a run share state --- transcript, memory store, observation buffer --- and are
not independent.

\subsection{Accuracy Benchmark Design}
\label{sec:acc-design}

\paragraph{Dataset.}
Accuracy is evaluated on \textbf{LoCoMo}, restricted to a
stratified subset --- four full dialogues, 665 evaluated QA pairs ---
chosen so its question-category mix matches the full corpus (\Cref{app:bench}).
Restricting accuracy evaluation to a single corpus anchors the joint
cost--accuracy analysis on one distribution.

\paragraph{Protocol.}
For each (system, setting) cell, the system ingests the full
conversation, retrieves from its memory at query time, and produces a
free-form or multiple-choice answer at $\text{temperature}{=}0.7$ with a
single pass per question.

\paragraph{Judge.}
The judge is \textbf{gpt-oss-120b} at $\text{temperature}{=}0$ with a single
pass per question. We hold it fixed across all setting cells so judging
is consistent across systems. Accuracy is the fraction of questions judged
correct, reported per (system, setting) cell.

\subsection{Joint Cost--Accuracy Framing}
\label{sec:joint-framing}

For each system at each setting we have a matched $(\hat{C}, \mathrm{Acc})$ pair,
where $\hat{C}$ is the mean billable cost at the $(N{=}100, L{=}100)$ reference
cell and $\mathrm{Acc}$ is the accuracy on the LoCoMo subset. We report this
joint matrix (\Cref{tab:joint}) and the cost-per-correct-answer ratio
$\hat{C}/\mathrm{Acc}$ as the primary derived statistic.

\section{Results}
\label{sec:results}

We report four things: (i)~the per-turn cost model and its fitted exponents for
the three memory systems and two baselines, (ii)~held-out cross-validation of
that model, (iii)~a cost break-even analysis of when a memory system becomes
cheaper to serve than the full-history baseline, and (iv)~accuracy on the
LoCoMo subset. We then present the joint cost--accuracy matrix.

\subsection{Per-Turn Cost Model}
\label{sec:loocv}

\Cref{tab:cost-model} reports the fitted exponents of the separable cost
model (\Cref{eq:multivariate}), its in-sample $R^2$, and held-out
LOOCV-MAPE. We summarize each as a range across the four (backbone,
reasoning-effort) settings. \Cref{tab:app-cost-model-full} gives the
per-(system, setting) breakdown.

\begin{table*}[t]
\centering\small
\begin{tabular}{l r r r r}
\toprule
System & $p$ ($L$) & $q$ ($t$) & $R^2$ & LOOCV-MAPE \\
\midrule
Full history & 0.95--0.97 & 0.94--0.97 & 0.996--0.999 & 0.029--0.053 \\
Rolling window ($k{=}10$)  & 0.85--0.92 & 0.10--0.12 & 0.913--0.950 & 0.061--0.065 \\
Mem0      & 0.16--0.18 & 0.07--0.09 & 0.048--0.069 & 0.184--0.222 \\
Hindsight & 0.14--0.17 & 0.41--0.43 & 0.320--0.348 & 0.461--0.478 \\
Mastra OM & 0.61--0.79 & 0.23--0.36 & 0.457--0.557 & 0.408--0.685 \\
\bottomrule
\end{tabular}
\caption{Separable per-turn cost model
$\log(C{+}1) = a + p\log(L{+}1) + q\log(t{+}1)$ fitted per (system, setting),
summarized as ranges across the four settings. $R^2$ is in-sample;
LOOCV-MAPE is leave-one-$(N,L)$-cell-out held-out error. Held-out folds per
row: full-history~8, rolling-window~8, Mastra OM~7, Mem0~5, Hindsight~5.
\Cref{tab:app-cost-model-full} gives the per-(system, setting) breakdown.}
\label{tab:cost-model}
\end{table*}

\paragraph{Fitted exponents.}
The two baselines set the reference points. The full-history baseline fits
$p \approx q \approx 1$ ($R^2 \geq 0.996$): every turn resubmits the whole
transcript, so cost is proportional to message size times depth.
The rolling-window baseline fits $p \approx 0.9$ but $q \approx 0.1$:
cost scales almost linearly with message size but is nearly flat in depth.
This is the result of a bounded ten-turn window. A single cumulative-content
predictor $T = N{\cdot}L$ cannot capture this depth--size split. Fit on
$\log(T{+}1)$ alone, the same rolling-window data reaches only
$R^2 = 0.39$--$0.40$, against the $0.91$--$0.95$ of the separable model. The
three memory systems show different cost patterns. Mem0 has small exponents on
both axes ($p {\approx} 0.17$, $q {\approx} 0.08$). Hindsight has a small
message-size exponent but the largest depth exponent ($p {\approx} 0.15$,
$q {\approx} 0.42$). Mastra OM has the largest message-size exponents of
the three ($p \in [0.61, 0.79]$, $q \in [0.23, 0.36]$) and the highest
in-sample $R^2$.

\paragraph{Held-out validation.}
LOOCV-MAPE separates the five configurations into two clear groups. The two
baselines generalize to within $2.9$--$6.5\%$. The memory systems do not:
$18$--$22\%$ for Mem0, $46$--$48\%$ for Hindsight, and $41$--$69\%$ for
Mastra OM. The worst cell is off by $1.9\times$ the true cost. The held-out
failures cluster at the grid corners: short conversations for Hindsight and
small-message ($L{=}50$) cells for Mastra OM. We return to this split in
\Cref{sec:discuss}.

\paragraph{Token-accounting diagnostics.}
\Cref{tab:app-cost-diag} reports three per-stage ratios that the cost model
does not capture on its own: the output-to-input token ratio $\gamma$, the
answer-stage reasoning-to-output ratio $\zeta_\mathrm{ans}$, and the ingest-stage
reasoning-to-output ratio $\zeta_\mathrm{ing}$. These ratios, not the cost
exponents, drive the cross-backbone cost gap discussed in
\Cref{sec:discuss}.

\subsection{Cost Break-Even: When a Memory System Pays for Itself}
\label{sec:breakeven}

A memory system pays for itself only once a conversation is long enough to
spread its overhead. We define the \emph{break-even length} as the turn at
which a system's cumulative cost drops below the full-history cost and stays
below it. We compute break-even from measured per-turn cost (8~reps averaged,
no fitted-model extrapolation). Each value therefore comes from real grid cells
and is not affected by the held-out prediction failure of \Cref{sec:loocv}.

\begin{figure}[t]
\centering
\includegraphics[width=\columnwidth]{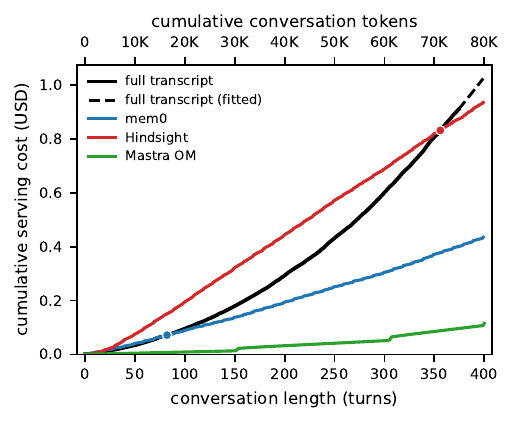}
\caption{Cumulative serving cost vs.\ conversation length for a 400-turn
conversation at 200 tokens per turn under \texttt{gpt-oss-20b} at low
reasoning effort (top axis:
cumulative conversation tokens). Dots mark the break-even turn at which a
system's cumulative cost drops below the full-history ceiling; the
full-history curve is measured to turn~374 and fitted (dashed) beyond (\Cref{sec:lim}).}
\Description{A line plot of cumulative serving cost (y-axis) against
conversation turn (x-axis, 0--400) for the three memory systems and the two
baselines on the gpt-oss-20b backbone at low reasoning effort. The full-history baseline grows superlinearly and dominates beyond
roughly 100 turns; Mastra OM stays lowest among the memory systems; Mem0 crosses the
full-history line near turn 82; Hindsight crosses near turn 356; the
rolling-window baseline is flat and lowest at all turns. Markers on each
memory-system curve indicate its break-even turn against full history.}
\label{fig:breakeven}
\end{figure}

\Cref{fig:breakeven} traces this for a 400-turn conversation under
\BackboneOss{}: Mastra OM breaks even at turn~0, Mem0 at turn~82,
and Hindsight only at turn~356. Across all measured 400-turn cells the
break-even length spans Mastra OM 0--86, Mem0 0--342, and Hindsight
60--never. By a 400-turn conversation the full transcript costs up to
$12.7\times$ a memory system that has broken even. Hindsight at small
messages, in turn, can cost up to $3.3\times$ the full transcript. Break-even
arrives earlier with larger per-turn messages and on \BackboneGemma{}. We
discuss these workload dependencies in \Cref{sec:discuss}.

\subsection{Accuracy on LoCoMo}
\label{sec:acc-results}

\Cref{tab:acc} reports per-(system, setting) accuracy on the 665
evaluated questions of the LoCoMo subset. Mem0 and Mastra OM cover a
wide range, $[0.21, 0.52]$, and both score higher on \BackboneGemma{} than on
\BackboneOss{}. We report Hindsight's per-cell accuracy here. But because its
ingest stage did not run under the per-cell backbone (\Cref{sec:lim}), we do
not compare its cells across backbones or reasoning levels.

\begin{table*}[t]
\centering\small
\setlength{\extrarowheight}{2pt}
\begin{tabular}{l r r r r}
\toprule
System & \SettingHeaderOssLow{} & \SettingHeaderOssMedium{} & \SettingHeaderGemmaLow{} & \SettingHeaderGemmaMedium{} \\
\midrule
Mem0      & \makecell[r]{0.322\\{\scriptsize[0.287,\,0.358]}} & \makecell[r]{0.214\\{\scriptsize[0.184,\,0.246]}} & \makecell[r]{0.498\\{\scriptsize[0.460,\,0.536]}} & \makecell[r]{0.516\\{\scriptsize[0.478,\,0.554]}} \\
Mastra OM & \makecell[r]{0.361\\{\scriptsize[0.325,\,0.398]}} & \makecell[r]{0.308\\{\scriptsize[0.274,\,0.344]}} & \makecell[r]{0.429\\{\scriptsize[0.391,\,0.466]}} & \makecell[r]{0.502\\{\scriptsize[0.464,\,0.540]}} \\
Hindsight$^\dagger$ & \makecell[r]{0.528\\{\scriptsize[0.490,\,0.565]}} & \makecell[r]{0.541\\{\scriptsize[0.503,\,0.579]}} & \makecell[r]{0.493\\{\scriptsize[0.455,\,0.531]}} & \makecell[r]{0.498\\{\scriptsize[0.460,\,0.536]}} \\
\bottomrule
\end{tabular}
\caption{Accuracy on the LoCoMo subset (665 evaluated questions),
per (system, setting) cell. Bracketed values are Wilson 95\%
confidence intervals~\cite{wilson1927probable,brown2001interval} at
$n{=}665$; they treat the 665 questions as
independent and so understate uncertainty given dialogue-level clustering
(\Cref{sec:lim}). $^\dagger$Hindsight's ingest stage ran under a configuration the benchmark did not control, so its cells are not
comparable across the backbone/reasoning columns (\Cref{sec:lim}).}
\label{tab:acc}
\end{table*}

\subsection{Joint Cost--Accuracy Matrix}
\label{sec:joint}

\Cref{tab:joint} pairs each cell's accuracy with its mean billable cost at
the reference cell ($N{=}100, L{=}100$), in USD per 100-turn conversation
(8 reps averaged), and with the cost-per-correct-answer ratio
$\hat{C}/\mathrm{Acc}$. We discuss the trade-off in \Cref{sec:discuss}.

\begin{table*}[t]
\centering\small
\begin{tabular}{l l r r r r}
\toprule
System & Metric & \SettingHeaderOssLow{} & \SettingHeaderOssMedium{} & \SettingHeaderGemmaLow{} & \SettingHeaderGemmaMedium{} \\
\midrule
\multirow{3}{*}{Mem0}
  & Acc                       & \makecell[r]{0.322\\{\scriptsize[0.287,\,0.358]}} & \makecell[r]{0.214\\{\scriptsize[0.184,\,0.246]}} & \makecell[r]{0.498\\{\scriptsize[0.460,\,0.536]}} & \makecell[r]{0.516\\{\scriptsize[0.478,\,0.554]}} \\
  & $\hat{C}$ (USD)           & \$0.059 & \$0.065 & \$0.019 & \$0.019 \\
  & $\hat{C}/\mathrm{Acc}$    & 0.183   & 0.305   & 0.038   & 0.037   \\
\midrule
\multirow{3}{*}{Mastra OM}
  & Acc                       & \makecell[r]{0.361\\{\scriptsize[0.325,\,0.398]}} & \makecell[r]{0.308\\{\scriptsize[0.274,\,0.344]}} & \makecell[r]{0.429\\{\scriptsize[0.391,\,0.466]}} & \makecell[r]{0.502\\{\scriptsize[0.464,\,0.540]}} \\
  & $\hat{C}$ (USD)           & \$0.010 & \$0.044 & \$0.030 & \$0.031 \\
  & $\hat{C}/\mathrm{Acc}$    & 0.028   & 0.143   & 0.070   & 0.062   \\
\midrule
\multirow{3}{*}{Hindsight$^\dagger$}
  & Acc                       & \makecell[r]{0.528\\{\scriptsize[0.490,\,0.565]}} & \makecell[r]{0.541\\{\scriptsize[0.503,\,0.579]}} & \makecell[r]{0.493\\{\scriptsize[0.455,\,0.531]}} & \makecell[r]{0.498\\{\scriptsize[0.460,\,0.536]}} \\
  & $\hat{C}$ (USD)           & \$0.240 & \$0.243 & \$0.051 & \$0.052 \\
  & $\hat{C}/\mathrm{Acc}$    & 0.455   & 0.450   & 0.104   & 0.104   \\
\bottomrule
\end{tabular}
\caption{Joint cost--accuracy matrix at the $(N{=}100, L{=}100)$ reference
cell. $\hat{C}$ is mean billable cost in USD per 100-turn conversation
over 8 reps, priced at the rates in \Cref{app:config}; Acc is from the
LoCoMo subset, with bracketed Wilson 95\% confidence intervals at $n{=}665$
(independent-question approximation; see the clustering caveat in
\Cref{sec:lim}). $^\dagger$Hindsight's ingest stage ran under a configuration the benchmark did not control, so its cells are not
comparable across the backbone/reasoning columns (\Cref{sec:lim}).}
\label{tab:joint}
\end{table*}

At the reference cell (\Cref{tab:joint}), most memory systems have not yet
paid back their ingest cost against full history. \Cref{tab:joint-400}
recomputes the matrix at $(N{=}400, L{=}200)$. There, all Mem0 and Mastra OM
cells, and all Hindsight cells except \SettingOssMedium{}, have crossed
break-even (\Cref{tab:app-breakeven}). Hindsight on \SettingOssMedium{}
stays slightly above the measured full-history cost (\$0.869 vs.\ \$0.858)
and is marked \emph{never}. Accuracy is measured once per cell on
the LoCoMo subset and does not depend on $(N, L)$. The Acc column and its
Wilson intervals are therefore identical to \Cref{tab:joint}; only $\hat{C}$
and the ratio change. The lowest-cost-per-correct configuration is unchanged:
Mastra OM on \SettingOssLow{} ($\hat{C}/\mathrm{Acc}{=}0.278$) and Mem0 on
\BackboneGemma{} ($0.325$--$0.339$). The lower half of the ranking, however,
flips. Hindsight is the costliest-per-correct system at $(100,100)$, but it is
no longer last in three of four columns. Mem0 on \SettingOssMedium{} now costs
more ($2.190$ vs.\ $1.607$), as does Mastra OM on both \BackboneGemma{} columns
($0.768$/$0.684$ vs.\ $0.543$/$0.542$). The reason is that Hindsight's large
fixed ingest cost spreads over a longer conversation, while Mem0 and Mastra OM
scale up faster ($3.4$--$5.3\times$ vs.\ $7$--$11\times$ from the reference
cell; these are $\hat{C}_{(400,200)}/\hat{C}_{(100,100)}$ ratios from
\Cref{tab:joint,tab:joint-400} for Hindsight vs.\ Mem0 and Mastra OM).
The full-history figures for \BackboneOss{} are measured only to turn~374
(\Cref{sec:lim}), so \BackboneOss{} comparisons should be read alongside the
break-even statuses in \Cref{tab:app-breakeven}.

\begin{table*}[t]
\centering\small
\begin{tabular}{l l r r r r}
\toprule
System & Metric & \SettingHeaderOssLow{} & \SettingHeaderOssMedium{} & \SettingHeaderGemmaLow{} & \SettingHeaderGemmaMedium{} \\
\midrule
\multirow{3}{*}{Mem0}
  & Acc                       & \makecell[r]{0.322\\{\scriptsize[0.287,\,0.358]}} & \makecell[r]{0.214\\{\scriptsize[0.184,\,0.246]}} & \makecell[r]{0.498\\{\scriptsize[0.460,\,0.536]}} & \makecell[r]{0.516\\{\scriptsize[0.478,\,0.554]}} \\
  & $\hat{C}$ (USD)           & \$0.435 & \$0.469 & \$0.169 & \$0.167 \\
  & $\hat{C}/\mathrm{Acc}$    & 1.350   & 2.190   & 0.339   & 0.325   \\
\midrule
\multirow{3}{*}{Mastra OM}
  & Acc                       & \makecell[r]{0.361\\{\scriptsize[0.325,\,0.398]}} & \makecell[r]{0.308\\{\scriptsize[0.274,\,0.344]}} & \makecell[r]{0.429\\{\scriptsize[0.391,\,0.466]}} & \makecell[r]{0.502\\{\scriptsize[0.464,\,0.540]}} \\
  & $\hat{C}$ (USD)           & \$0.100 & \$0.301 & \$0.330 & \$0.344 \\
  & $\hat{C}/\mathrm{Acc}$    & 0.278   & 0.979   & 0.768   & 0.684   \\
\midrule
\multirow{3}{*}{Hindsight$^\dagger$}
  & Acc                       & \makecell[r]{0.528\\{\scriptsize[0.490,\,0.565]}} & \makecell[r]{0.541\\{\scriptsize[0.503,\,0.579]}} & \makecell[r]{0.493\\{\scriptsize[0.455,\,0.531]}} & \makecell[r]{0.498\\{\scriptsize[0.460,\,0.536]}} \\
  & $\hat{C}$ (USD)           & \$0.822 & \$0.869 & \$0.268 & \$0.270 \\
  & $\hat{C}/\mathrm{Acc}$    & 1.557   & 1.607   & 0.543   & 0.542   \\
\bottomrule
\end{tabular}
\caption{Joint cost--accuracy matrix at the $(N{=}400, L{=}200)$ cell; all
Mem0 and Mastra OM cells, and all Hindsight cells except \SettingOssMedium{},
have crossed break-even against full history (\Cref{tab:app-breakeven}).
$\hat{C}$ is mean billable cost in USD per 400-turn conversation over 8 reps,
priced with the same accounting as \Cref{tab:joint} (rates in
\Cref{app:config}); Acc is identical to \Cref{tab:joint} because it is measured
once per cell on the LoCoMo subset and is independent of $(N, L)$, with
bracketed Wilson 95\% confidence intervals at $n{=}665$ (independent-question
approximation; see the clustering caveat in \Cref{sec:lim}).
$^\dagger$Hindsight's ingest stage ran under a configuration the benchmark did not control, so its cells are not comparable across the
backbone/reasoning columns (\Cref{sec:lim}).}
\label{tab:joint-400}
\end{table*}

\section{Discussion}
\label{sec:discuss}

\subsection{Separating Conversation Depth from Message Size}

A single cumulative-content predictor $T = N{\cdot}L$ treats two different
cases as the same: a long conversation of small messages and a short
conversation of large messages. Memory systems handle these two cases
differently. We therefore fit message size $L$ and conversation depth $t$ as
separate exponents (\Cref{tab:cost-model}). For the window-based baselines the
exponents match the known mechanism. The full-history baseline fits $p \approx
q \approx 1$, because every token is resubmitted. The rolling-window baseline
fits $q \approx 0.11$, so cost is nearly flat in depth; this follows from its
bounded window. Held-out error for both stays below $6.5\%$. Using two
predictors instead of one captures this structure: the rolling-window in-sample
$R^2$ rises from $0.39$--$0.40$ under a one-variable power law to $0.91$--$0.95$ here.

The same separable model does \emph{not} hold out for the memory systems.
A low in-sample $R^2$ on its own does not show that a system has separated cost
from conversation size. Mastra OM has the highest in-sample $R^2$ of the three
memory systems, yet it has the worst held-out error. The held-out test shows
that per-turn cost depends on internal memory state. To predict serving cost for
these systems at an unseen workload, we would need a model of the memory
subsystem itself. We leave this to future work (\Cref{sec:lim}).

\subsection{Backbone and Reasoning Effort Shape the Cost Surface}

The backbone's effect on cost depends on the system and the reasoning level. It
does not favor one model in every case. Mem0 shows this most plainly. Its
reference-cell cost falls from \$0.059--\$0.065 under \BackboneOss{} to
\$0.019 under \BackboneGemma{}, and its $(N{=}400,L{=}200)$ cost falls from
\$0.435--\$0.469 to \$0.167--\$0.169 (\Cref{tab:joint,tab:joint-400}). This
gap matches the diagnostics in \Cref{tab:app-cost-diag}. Mem0's
answer- and ingest-stage reasoning-to-output ratios are near one on
\BackboneOss{} but zero on \BackboneGemma{}, and its output-to-input
ratio $\gamma$ exceeds one only on \BackboneOss{}. Mastra OM behaves in the
opposite way. Its \SettingOssLow{} cell is the cheapest Mastra OM setting in
both joint matrices, which fits its lower $\gamma$ and lower reasoning ratios
than its \BackboneGemma{} cells. We exclude Hindsight from cross-backbone
comparisons because its ingest stage was not benchmark-controlled
(\Cref{sec:lim}). In practice, backbone and memory choice are not separable: 
whether a given backbone reduces or increases cost, and by how much, is governed 
by the memory system's internal call pattern.

Raising reasoning effort from \emph{low} to \emph{medium} increases cost but
does not always improve accuracy. Mem0's accuracy on \BackboneOss{}
\emph{drops} from $0.322$ to $0.214$ at the higher reasoning level. This is
likely because reasoning tokens use up the \texttt{max\_tokens} budget and
leave less room for the answer. Mastra OM moves in opposite directions on the
two backbones ($+7.3$~pp on \BackboneGemma{}, $-5.3$~pp on \BackboneOss{}).
Reasoning effort is therefore not a simple cost knob: at the same step, the
accuracy change varies by $\geq 10$~percentage points across systems.

\subsection{When Does a Memory System Pay Off?}

No single system is best in all settings. Mem0's extraction-based ingest keeps
its per-turn cost the flattest across $(L, t)$, but its accuracy varies the
most of the three (from $0.214$ to $0.516$). Hindsight is the most
expensive system to serve: $\hat{C} \approx 0.24$~USD per 100-turn
conversation under \BackboneOss{}, several times any other cell. Mastra OM's
threshold-based reflector fires based on internal memory state, so its
per-turn cost does not follow conversation length or message size
(\Cref{sec:loocv}). The lowest cost-per-correct-answer is Mastra OM on
\SettingOssLow{} ($0.028$ at $(100,100)$ and $0.278$ at $(400,200)$). Mem0 is
the cheapest option on \BackboneGemma{} ($0.037$--$0.038$ at
$(100,100)$ and $0.325$--$0.339$ at $(400,200)$). Hindsight on \BackboneOss{}
is the most expensive at the reference cell ($\hat{C}/\mathrm{Acc} \approx 0.45$).

The right choice depends on the workload. A memory system is cheaper than
full-history serving only after it crosses its break-even length
(\Cref{sec:breakeven}). Mastra OM breaks even at once, Mem0 within tens of
turns once messages are large, and Hindsight only late, sometimes past 400
turns. Whether a memory system is worth its cost therefore depends on the
expected conversation length and the chosen backbone, not on the system alone.
Full-history cost grows without limit ($p \approx q \approx 1$,
\Cref{sec:loocv}) and must eventually fill any finite context window. Within
our 400-turn grid, however, no overflow occurs (\Cref{sec:lim}).

\subsection{Limitations}
\label{sec:lim}

\paragraph{Synthetic cost-benchmark dialogues.}
The cost benchmark uses LLM-generated dialogues rather than real conversations.
This choice is deliberate: billing depends on token counts, which are set
by message size and conversation length. Ingest-side behavior, however, can
depend on content. Memory fact extraction and memory consolidation rates depend
on how many extractable facts appear per turn, and this density may differ in
real conversations.

\paragraph{Single accuracy corpus.}
Accuracy is reported on the LoCoMo subset only. The joint claims hold for
the LoCoMo distribution (open-domain persona-grounded multi-session chat)
and do not directly transfer to task-oriented or knowledge-intensive QA.
Validating the fitted cost model on a task-oriented corpus
(MultiWOZ~2.2~\cite{zang-etal-2020-multiwoz}) is planned as future work.

\paragraph{Cost model is descriptive, not mechanistic.}
The cost model in \Cref{eq:multivariate} is a regression on conversation
length and message size. It generalizes well to the two window-based baselines,
but performs poorly on Mem0, Hindsight, and Mastra OM, where leave-one-cell-out
errors range from $0.18$ to $0.69$ (\Cref{tab:cost-model}, \Cref{sec:discuss}).
Cost predictions for these three systems at untested $(N, L)$ values should be
read as rough estimates only. We leave a mechanistic model that tracks internal
memory state for each system to future work.

\paragraph{Hindsight's ingest backbone was not benchmark-controlled.}
Hindsight's ingest stage (fact extraction, consolidation, recall ranking)
runs in an external self-hosted HTTP server. Its base environment file overrode the
per-cell backbone we set, so ingest ran under a single fixed configuration
(\BackboneOss{} at a fixed reasoning effort) across all four settings.
Only the answer stage, which the harness sets directly, followed the grid.
We report Hindsight's per-cell totals
(\Cref{tab:app-cost-model-full,tab:app-cost-diag,tab:acc,tab:joint}) but
exclude it from cross-backbone and reasoning-effort comparisons and from
$\zeta_\mathrm{ing}$. Mem0 and Mastra OM ingest in-process and follow the
per-cell configuration (\Cref{tab:app-cost-diag}).

\paragraph{Provider-routing variance.}
OpenRouter's provider-preference list ($\langle$\,groq,
amazon-bedrock, google-vertex\,$\rangle$) holds across our runs, but the
primary provider can saturate and route requests to a fallback provider.
This affects cost in two ways. First, per-token rates differ across
providers, so observed billing can differ by single-digit percent from the
contract rates used in our fitted costs. Second, prompt caching is tied to
one provider. A prefix cached at the primary provider does not hit on the
fallback provider, so its tokens are billed at the full input rate
($\$0.075$/M) instead of the cached rate ($\$0.0375$/M, half price).
Systems that resend a large fixed context with each answer are most affected.
For example, Mastra OM builds up to $40{,}000$ observation tokens.
These tokens hit the cached rate on the same provider, but cost twice as
much on the input segment when a fallback breaks the cache.

\paragraph{Cost-per-correct-answer is a partial quality metric.}
The statistic $\hat{C}/\mathrm{Acc}$ in \Cref{tab:joint} ranks systems by
cost per LoCoMo-judged correct answer. It treats accuracy as the only quality
measure and ignores latency, retrieval-payload size, answer-token budget,
retrieval recall, and abstention behavior, any of which can matter in
practice. We report the ratio because it follows from the matched
cost--accuracy design. We do not claim that minimizing it is the right
deployment goal.

\paragraph{Serving-stack token-count mismatch.}
On \BackboneOss{}, the OpenRouter\,$\rightarrow$\,\allowbreak groq serving stack rejected
full-history answer-stage requests beyond turn~374 of the $(N{=}400, L{=}200)$
cell. The rejection cited a 98{,}516-token prompt, but the true prompt is
about 74{,}800 tokens and fits within the 131{,}072-token context
window. \BackboneGemma{} completed all 400 turns of the same conversation.
We treat this as a serving-stack error, report \BackboneOss{} full-history
costs only to turn~374, and make no context-length claim. \Cref{fig:breakeven}
extends the curve using the fitted model (held-out error below $6.5\%$,
\Cref{sec:loocv}).

\section{Conclusion}
\label{sec:conclusion}

This study measures the cost and accuracy of three memory systems under
a shared benchmark. We fit a separable cost model
($\log(C{+}1) = a + p\,\log(L{+}1) + q\,\log(t{+}1)$)
and validate it with leave-one-cell-out cross-validation. Accuracy is scored
by a fixed judge on the LoCoMo subset. The main cost finding is that the
model predicts well for window-based strategies but not for memory systems.
Full-history ($p \approx q \approx 1$) and rolling-window ($q \approx 0$) both
hold out below $6.5\%$ error. The retrieval- and threshold-driven memory
systems have $18$--$69\%$ held-out error. This shows that their cost is
driven by internal memory state, not by conversation length or message size.

No single system is best in all settings. Mem0's accuracy varies the most
($0.214$--$0.516$). Hindsight is the most expensive at the 100-turn reference
cell. The lowest cost-per-correct-answer is Mastra OM on \SettingOssLow{}
($0.028$ at $(100,100)$ and $0.278$ at $(400,200)$), while Mem0 leads on
\BackboneGemma{} ($0.037$--$0.038$ at $(100,100)$ and $0.325$--$0.339$ at
$(400,200)$). The backbone choice (\BackboneGemma{} vs.\ \BackboneOss{})
shapes the cost surface as much as the system choice, so backbone and
memory are a joint decision. Finally, a break-even analysis shows that a memory
system becomes cheaper than full-history serving only past a threshold that
depends on the workload. This threshold is immediate for the cheapest systems
and beyond 400 turns for the most expensive. The choice therefore depends on
the expected conversation length, not on the memory system alone.

\appendix
\crefalias{section}{appendix}

\section{System and Backbone Configurations}
\label{app:config}

\Cref{tab:app-systems} lists the per-system configuration --- the ingest
trigger, retrieval setting, and the memory-specific parameters the main text names 
but does not specify values for. Both backbones share temperature $0.7$,
\texttt{max\_tokens} $32{,}768$, and per-cell reasoning effort with provider
fallbacks disabled, but resolve to different OpenRouter provider stacks:
$\langle$groq, amazon-bedrock, google-vertex$\rangle$ for
\BackboneOss{} and $\langle$deepinfra/fp8, io-net/bf16, cloudflare$\rangle$
for \BackboneGemma{}. The judge (\texttt{gpt-oss-120b}) and the dialogue generator
(\texttt{gemma-4-26b-a4b-it}, \texttt{max\_tokens} $8{,}192$) are held fixed
across runs. Per-token pricing (\$/M tokens) varies by provider. For
\texttt{gpt-oss-20b}: groq \$0.075 (input), \$0.0375 (cached input), \$0.30
(output); amazon-bedrock \$0.07 (input), \$0.15 (output); google-vertex \$0.07
(input), \$0.25 (output). For \texttt{gemma-4-26b-a4b-it}: deepinfra/fp8 \$0.07
(input), \$0.34 (output); io-net/bf16 \$0.15 (input), \$0.15 (cached input),
\$0.50 (output); cloudflare \$0.10 (input), \$0.30 (output). Embeddings use
\texttt{pplx-embed-v1-0.6b} at \$0.004 (embedding input).

\begin{table*}[t]
\centering\small
\begin{tabular}{l l l p{0.42\textwidth}}
\toprule
System & Ingest trigger & Retrieval & Remarks \\
\midrule
Mem0 & every 10 turns & top-$k{=}10$ & -- \\
Hindsight & every 10 turns & top-$k{=}10$ &
Ingest-stage LLM is \emph{not} benchmark-controlled; recall budget \emph{mid}, max\_tokens=4{,}096. \\
Mastra OM & token threshold & top-$k{=}1$ over observations &
observer is triggered at 30{,}000 accumulated message
tokens, reflector at 40{,}000 accumulated observation tokens. \\
Full history & every turn & --- (full transcript) &
-- \\
Rolling window ($k{=}10$) & every turn & --- (last 10 turns) &
-- \\
\bottomrule
\end{tabular}
\caption{Per-system configuration. The benchmark controls the answer-stage LLM
for all systems and the ingest-stage LLM for Mem0 and Mastra OM; Hindsight's
ingest LLM is not benchmark-controlled (\Cref{sec:lim}).}
\label{tab:app-systems}
\end{table*}

\section{Benchmark Construction and Fitting Procedures}
\label{app:bench}

\paragraph{Cost grid.}
We fit the memory systems on five cells: the four corners and the center,
$(N, L) \in \{(15,50)$, $(15,200)$, $(100,100)$, $(400,50)$, $(400,200)\}$, at 8 reps
each. Mastra OM adds two cells, $(800,200)$ and $(800,400)$, at 3 reps to reach
its threshold regime. The two deterministic baselines use eight cells at 2 reps:
the five above plus three matched-content cells, $(40,250), (200,50), (400,25)$.
Each cost fit uses every turn of every run: 40 runs for Mem0 and Hindsight,
46 for Mastra OM, and 16 per baseline.

\paragraph{Synthetic dialogue generation.}
For each $(N, L, \text{seed})$, the generator runs in chunks of 20 turns, using
the chunk prompt in \Cref{app:prompts}. We tokenize each turn with
\texttt{o200k\_base}~\cite{tiktoken}. Turns above $1.15{\cdot}L$ are trimmed and
turns below $0.85{\cdot}L$ are padded with filler words (tolerance $0.15$). We
cache each completed dialogue on disk under its $(N, L, \text{seed})$ key, so
the conversation is the same across reps and across systems.

\paragraph{LoCoMo subset selection.}
We score candidate four-dialogue subsets by the Jensen--Shannon
divergence~\cite{MENENDEZ1997307} between the subset and the full dataset,
measured across question category, evidence count, and evidence span. We keep
the subset with the lowest total divergence ($<3\cdot10^{-5}$ per axis).
Following dataset convention, we exclude Category-5 adversarial questions, which
leaves 665 evaluated QA pairs.

\paragraph{Cost-model fitting.}
We compute one OLS fit of \Cref{eq:multivariate} per (system, setting), with
$\log(C{+}1)$ as the response. The 95\% confidence intervals for the exponents
come from a cluster bootstrap that resamples whole runs with replacement
($B{=}1{,}000$). LOOCV-MAPE holds out one $(N, L)$ cell at a time, refits on the
remaining cells, and predicts the held-out cell's mean per-turn cost. The number
of folds per system equals the number of cells (8 for the baselines, 7 for
Mastra OM, 5 for Mem0 and Hindsight). For the memory systems, the held-out
errors cluster at the short-conversation and small-message grid corners
(\Cref{sec:loocv}). \Cref{tab:app-cost-model-full} gives the per-(system,
setting) fits that \Cref{tab:cost-model} summarizes, and \Cref{tab:app-cost-diag}
gives the token-accounting diagnostics cited in \Cref{sec:loocv} and
\Cref{sec:discuss}.

\begin{table*}[t]
\centering\small
\begin{tabular}{l l r r r r r}
\toprule
System & Setting & $p$ ($L$) & $q$ ($t$) & $R^2$ & LOOCV-MAPE & max APE \\
\midrule
Full history & \SettingGemmaLow{}    & 0.97 & 0.97 & 0.999 & 0.029 & 0.089 \\
Full history & \SettingGemmaMedium{} & 0.97 & 0.97 & 0.999 & 0.030 & 0.089 \\
Full history & \SettingOssLow{}      & 0.95 & 0.94 & 0.996 & 0.053 & 0.145 \\
Full history & \SettingOssMedium{}   & 0.95 & 0.94 & 0.996 & 0.053 & 0.145 \\
Rolling window ($k{=}10$)  & \SettingGemmaLow{}    & 0.91 & 0.11 & 0.913 & 0.062 & 0.148 \\
Rolling window ($k{=}10$)  & \SettingGemmaMedium{} & 0.92 & 0.12 & 0.945 & 0.061 & 0.142 \\
Rolling window ($k{=}10$)  & \SettingOssLow{}      & 0.85 & 0.10 & 0.949 & 0.065 & 0.128 \\
Rolling window ($k{=}10$)  & \SettingOssMedium{}   & 0.85 & 0.10 & 0.950 & 0.065 & 0.128 \\
Mem0      & \SettingGemmaLow{}    & 0.18 & 0.08 & 0.069 & 0.184 & 0.244 \\
Mem0      & \SettingGemmaMedium{} & 0.16 & 0.09 & 0.067 & 0.185 & 0.254 \\
Mem0      & \SettingOssLow{}      & 0.16 & 0.07 & 0.048 & 0.222 & 0.282 \\
Mem0      & \SettingOssMedium{}   & 0.18 & 0.08 & 0.064 & 0.200 & 0.243 \\
Hindsight & \SettingGemmaLow{}    & 0.14 & 0.43 & 0.348 & 0.477 & 0.555 \\
Hindsight & \SettingGemmaMedium{} & 0.15 & 0.43 & 0.345 & 0.478 & 0.558 \\
Hindsight & \SettingOssLow{}      & 0.15 & 0.41 & 0.328 & 0.476 & 0.569 \\
Hindsight & \SettingOssMedium{}   & 0.17 & 0.41 & 0.320 & 0.461 & 0.570 \\
Mastra OM & \SettingGemmaLow{}    & 0.61 & 0.23 & 0.457 & 0.408 & 0.820 \\
Mastra OM & \SettingGemmaMedium{} & 0.68 & 0.25 & 0.458 & 0.685 & 1.918 \\
Mastra OM & \SettingOssLow{}      & 0.79 & 0.36 & 0.557 & 0.477 & 0.864 \\
Mastra OM & \SettingOssMedium{}   & 0.79 & 0.34 & 0.520 & 0.565 & 1.596 \\
\bottomrule
\end{tabular}
\caption{Per-(system, setting) fits of the separable cost model
$\log(C{+}1) = a + p\log(L{+}1) + q\log(t{+}1)$. $R^2$ is in-sample;
LOOCV-MAPE and max APE are leave-one-$(N,L)$-cell-out held-out errors.
Full-history holds out 8 folds per row, rolling-window 8, Mastra OM 7,
Mem0 5, Hindsight 5 (all five $(N, L)$ cells are completed for Hindsight
under every setting). \Cref{tab:cost-model} summarizes these rows
as ranges per system.}
\label{tab:app-cost-model-full}
\end{table*}

\begin{table}[!htbp]
\centering\small
\begin{tabular}{l l r r r}
\toprule
System & Setting & $\gamma$ & $\zeta_\mathrm{ans}$ & $\zeta_\mathrm{ing}$ \\
\midrule
Hindsight & \SettingCellGemmaLow{}    & 0.16 & 0.00 & -- \\
Hindsight & \SettingCellGemmaMedium{} & 0.16 & 0.00 & -- \\
Hindsight & \SettingCellOssLow{}      & 0.16 & 1.01 & -- \\
Hindsight & \SettingCellOssMedium{}   & 0.16 & 1.00 & -- \\
Mem0      & \SettingCellGemmaLow{}    & 0.56 & 0.00 & 0.00 \\
Mem0      & \SettingCellGemmaMedium{} & 0.55 & 0.00 & 0.00 \\
Mem0      & \SettingCellOssLow{}      & 1.27 & 0.99 & 0.93 \\
Mem0      & \SettingCellOssMedium{}   & 1.40 & 1.01 & 0.89 \\
Mastra OM & \SettingCellGemmaLow{}    & 0.43 & 1.03 & 0.83 \\
Mastra OM & \SettingCellGemmaMedium{} & 0.44 & 1.02 & 0.81 \\
Mastra OM & \SettingCellOssLow{}      & 0.23 & 0.36 & 0.31 \\
Mastra OM & \SettingCellOssMedium{}   & 0.53 & 0.96 & 0.74 \\
\bottomrule
\end{tabular}
\caption{Per-stage token-accounting diagnostics: output-to-input token ratio
$\gamma$, answer-stage reasoning-to-output ratio $\zeta_\mathrm{ans}$, and
ingest-stage reasoning-to-output ratio $\zeta_\mathrm{ing}$, computed from the
per-cell run logs. Each $\zeta$ is the ratio of reasoning (thinking)
tokens to visible-response tokens reported by the API; values above 1 occur
when the model spends more tokens on internal reasoning than on its visible
response, which is possible when the API reports them as separate counts.
The two baselines incur no ingest-stage LLM cost and are omitted.
Hindsight's $\zeta_\mathrm{ing}$ is omitted (--): its ingest-stage LLM ran
under a configuration the benchmark did not control (\Cref{sec:lim}).}
\label{tab:app-cost-diag}
\end{table}

\paragraph{Break-even computation.}
We compute the break-even turn from measured per-turn cost (8~reps averaged),
with no fitted-model extrapolation. For each system and for the full-history
baseline, we add up per-turn cost over depth. The break-even turn is the
smallest $t$ at which the system's cumulative cost falls below full-history's
and \emph{stays} below it at every later turn (\emph{immediate} $=$ cheaper from
turn~0; \emph{never} $=$ still costlier at the last measured turn).
\Cref{tab:app-breakeven} reports it for all measured multi-system cells. For
\BackboneOss{} full-history at $(N{=}400, L{=}200)$, the run ended at
turn~374 because of a serving-stack error (\Cref{sec:lim}), so we report
full-history cost only to that turn.

\begin{table}[t]
\centering\small
\begin{tabular}{l l c c c}
\toprule
Setting & Cell ($N{\times}L$) & Mem0 & Hindsight & Mastra OM \\
\midrule
\multirow{3}{*}{\SettingCellGemmaLow{}}
 & 100$\times$100 & 10  & never & 54 \\
 & 400$\times$50  & 23  & 223   & 86 \\
 & 400$\times$200 & 0   & 60    & 50 \\
\midrule
\multirow{3}{*}{\SettingCellGemmaMedium{}}
 & 100$\times$100 & 0   & never & 54 \\
 & 400$\times$50  & 21  & 224   & 73 \\
 & 400$\times$200 & 0   & 61    & 68 \\
\midrule
\multirow{3}{*}{\SettingCellOssLow{}}
 & 100$\times$100 & 82  & never & 0 \\
 & 400$\times$50  & 260 & never & 0 \\
 & 400$\times$200 & 82  & 356   & 0 \\
\midrule
\multirow{3}{*}{\SettingCellOssMedium{}}
 & 100$\times$100 & never & never & 13 \\
 & 400$\times$50  & 342   & never & 62 \\
 & 400$\times$200 & 40    & never & 28 \\
\bottomrule
\end{tabular}
\caption{Sustained break-even turn per (setting, cell, system).
\emph{never} = the system is still costlier than the full transcript at the
last measured turn. By turn~400 the full transcript costs up to $12.7\times$ a
system that has broken even (maximum ratio of full-history to adapter
cumulative cost at the last measured turn, over all setting/cell/adapter
combinations where the adapter has crossed break-even; attained by Mastra OM
on \SettingOssLow{} at the $400{\times}50$ cell). Hindsight at small messages
costs up to $3.3\times$ the full transcript (maximum ratio of Hindsight to
full-history cumulative cost over all 400-turn \emph{never} cells; attained on
\SettingOssMedium{} at the $400{\times}50$ cell).}
\label{tab:app-breakeven}
\end{table}

\section{Prompt Templates}
\label{app:prompts}

The two boxes below give the LLM-as-judge and dialogue-generation prompts
verbatim. Answer-stage prompts are system-specific --- each instructs a brief
answer grounded only in the retrieved memory or transcript, with relative time
references resolved to absolute dates from message timestamps. We used each
system's default prompt without modification; readers should refer to the
respective cited papers and system documentation for those templates.

\begin{tcolorbox}[breakable, colback=gray!5, colframe=gray!40, title=LLM-as-Judge Prompt (\texttt{gpt-oss-120b})]
\footnotesize
\textbf{System.} You are an expert grader that determines if answers to
questions match a gold standard answer.

\textbf{User.} Your task is to label an answer to a question as `CORRECT' or `WRONG'. You will
be given the following data: (1) a question (posed by one user to another
user), (2) a `gold' (ground truth) answer, (3) a generated answer, which you
will score as CORRECT/WRONG.

The point of the question is to ask about something one user should know about
the other user based on their prior conversations. The gold answer will usually
be a concise and short answer that includes the referenced topic. The generated
answer might be much longer, but you should be generous with your grading --- as
long as it touches on the same topic as the gold answer, it should be counted
as CORRECT.

For time-related questions, the gold answer will be a specific date, month,
year, etc. The generated answer might be much longer or use relative time
references (e.g.\ ``last Tuesday'' or ``next month''), but you should be generous
with your grading --- as long as it refers to the same date or time period as
the gold answer, it should be counted as CORRECT. Even if the format differs
(e.g.\ ``May 7th'' vs.\ ``7 May''), consider it CORRECT if it is the same date.

\textbf{Question:} \texttt{\{question\}}\quad
\textbf{Gold answer:} \texttt{\{golden\_answer\}}\quad
\textbf{Generated answer:} \texttt{\{generated\_answer\}}

First, provide a short (one sentence) explanation of your reasoning, then finish
with CORRECT or WRONG. Do NOT include both CORRECT and WRONG in your response.
Return the label in JSON format with the key \texttt{"label"}.
\end{tcolorbox}

\begin{tcolorbox}[breakable, colback=gray!5, colframe=gray!40, title=Synthetic Dialogue Generation --- Chunk Prompt]
\footnotesize
Generate \texttt{\{chunk\_size\}} turns of a natural two-speaker personal
conversation. Each turn should be about \texttt{\{l\_tokens\}} raw tokens long
under the o200k\_base tokenizer. Keep facts concrete: names, preferences,
routines, dates, plans, relationships. Return JSON only, as an object with key
\texttt{"turns"} whose value is an array of strings. No markdown.
Seed: \texttt{\{seed\}}. First turn index: \texttt{\{start\_idx\}}.
\end{tcolorbox}

\section*{Disclosure of Generative AI Usage}
In preparing this work, the authors used Claude Code (Opus 4.7 and
Sonnet 4.6) as an assistive tool. The research questions, the benchmark
design, the choice of memory systems, backbones, and evaluation metrics,
and the analysis and interpretation of the results represent the authors'
own novel intellectual contributions. Within that direction, the tool was
used to (i) support brainstorming and refinement of framing; (ii) write
and debug the benchmarking and analysis code that produced the results;
and (iii) draft and revise the text of the manuscript, including tables
and figure captions. The authors reviewed and verified all generated
content---confirming the correctness of the code, the validity of the
experimental results, and the accuracy of all claims and citations---and
edited the text as needed. The authors take full responsibility for the
veracity and correctness of the entire content of this work.

\clearpage
\bibliographystyle{ACM-Reference-Format}
\bibliography{custom}

\end{document}